\documentclass[runningheads]{llncs}
\usepackage{graphicx}
\usepackage{comment}
\usepackage{amsmath,amssymb} 
\usepackage{color}

\usepackage{multirow}
\usepackage[font=small]{caption}
\usepackage{epsfig}
\usepackage{subcaption}
\usepackage{amsfonts, float}
\usepackage{booktabs}
\usepackage{caption}
\usepackage{xcolor}
\graphicspath{ {images/} }
\usepackage[normalem]{ulem}

\usepackage[pagebackref=true,breaklinks=true,letterpaper=true,colorlinks,bookmarks=false]{hyperref}

\usepackage{multirow}
\usepackage{systeme,mathtools,enumerate}

\newcommand\Mark[1]{\textsuperscript#1}

\usepackage{authblk}
\begin{document}

\title{Fashion Captioning: Towards Generating Accurate Descriptions with Semantic Rewards} 

\titlerunning{Fashion Captioning}
\authorrunning{Xuewen Yang, \textit{et al.}}
\author{Xuewen Yang\inst{1} \and
Heming Zhang\inst{2} \and
Di Jin\inst{3} \and 
Yingru Liu\inst{1} \and
Chi-Hao Wu\inst{2} \and
Jianchao Tan\inst{4} \and
Dongliang Xie\inst{5} \and
Jue Wang\inst{6} \and
Xin Wang\inst{1}}
\institute{\Mark{1}Stony Brook University
\\ \email{xuewen.yang@stonybrook.edu} \\
\Mark{2}USC \quad \quad
\Mark{3}MIT \quad \quad  \Mark{4}Kwai Inc. \quad \quad
\Mark{5}BUPT \quad \quad
\Mark{6}Megvii
}

\maketitle

\begin{abstract}
Generating accurate descriptions for online fashion items is important not only for enhancing customers' shopping experiences, but also for the increase of online sales. 
Besides the need of correctly presenting the attributes of items, the expressions in an enchanting style could better attract customer interests.
The goal of this work is to develop a novel learning framework for accurate and expressive fashion captioning. 
Different from popular work on image captioning, it is hard to identify and describe the rich attributes of fashion items.
We seed the description of an item by first identifying its attributes, and introduce
\textit{attribute-level semantic} (ALS) reward and \textit{sentence-level semantic} (SLS) reward as metrics to improve the quality of text descriptions. 
We further integrate the training of our model with maximum likelihood estimation (MLE), attribute embedding, and Reinforcement Learning (RL).
To facilitate the learning,  we build a new FAshion CAptioning Dataset (FACAD), which contains $993$K images and $130$K corresponding enchanting and diverse descriptions.
Experiments on FACAD demonstrate the effectiveness of our model.\footnote{Code and data: \url{https://github.com/xuewyang/Fashion_Captioning}.}
\end{abstract}
\section{Introduction}
Motivated by the quick global growth of the fashion industry, which is worth trillions of dollars,
 extensive efforts have been devoted to fashion related research over the last few years. Those research directions include clothing attribute prediction and landmark detection \cite{Wang2018AttentiveFG,liuLQWTcvpr16DeepFashion}, fashion recommendation \cite{Yu2018}, item retrieval \cite{Liu:2012,Wang2017ClothingRW}, clothing parsing \cite{2018arXiv180610787G,He2017RealTimeFC}, and outfit recommendation \cite{han2017learning,Lu_2019_CVPR,Cucurull_2019_CVPR}.

\begin{figure}
\centering
\includegraphics[width=0.7\textwidth]{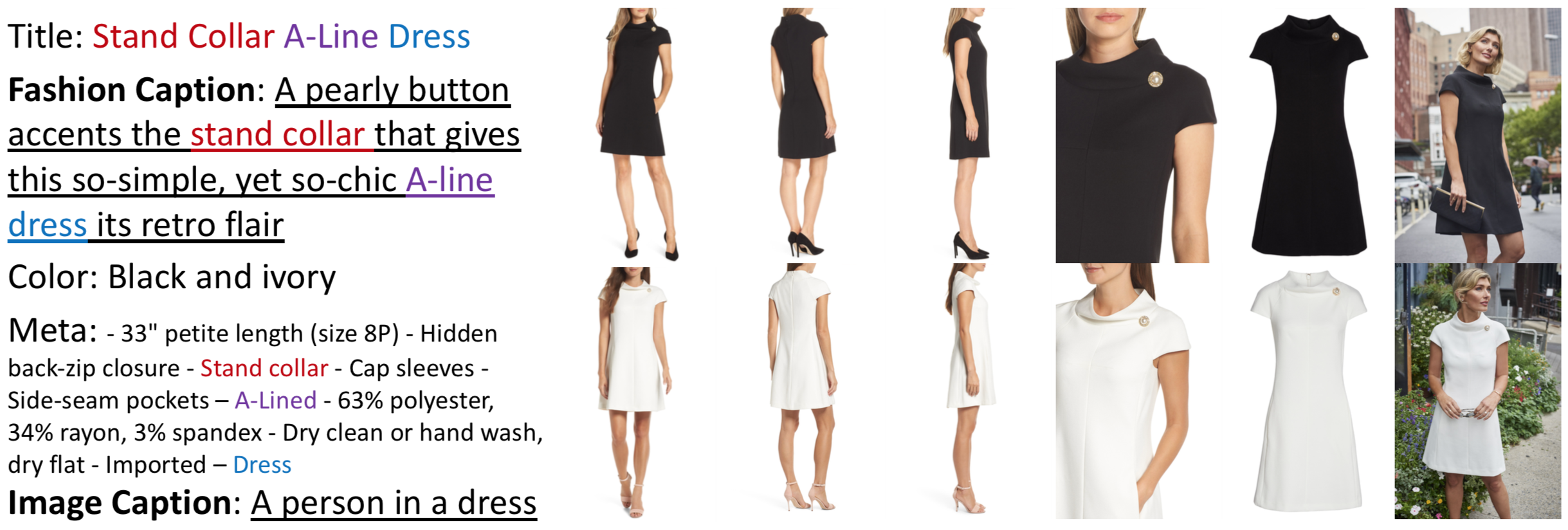}
\caption{An example for Fashion Captioning. The images are of different perspectives, colors and scenarios (shop-street). Other information contained include a title, a description (caption) from a fashion expert, the color info and the meta info. Words in color denotes the attributes used in sentence.
}
\label{fig:sample}
\end{figure}

Accurate and enchanting descriptions of clothes on shopping websites can help customers without fashion knowledge to better understand the features (attributes, style, functionality, benefits to buy, etc.) of the items and increase online sales  by enticing more customers.
However, manually writing the descriptions is a non-trivial and highly expensive task. 
Thus, the automatic generation of descriptions is in urgent need.
Since there exist no studies on generating fashion related descriptions, in this paper, we propose specific schemes on \textit{Fashion Captioning}. Our design is built upon our newly created \textit{FAshion CAptioning Dataset} (FACAD), the first fashion captioning dataset consisting of over $993$K images and $130$K descriptions with massive attributes and categories.
Compared with general image captioning datasets (e.g. MS COCO~\cite{ChenCOCO15}), 
the descriptions of fashion items have three unique features (as can be seen from Fig.~\ref{fig:sample}), which makes the automatic generation of captions a challenging task. 
First, fashion captioning needs to describe the fine-grained attributes of a single item, while image captioning generally narrates the objects and their relations in the image (e.g., a person in a dress). 
Second, the expressions to describe the clothes tend to be long so as to present the rich attributes of fashion items. The average length of captions in FACAD is 21 words while a sentence in the MS COCO caption dataset contains 10.4 words in average.
Third, FACAD has a more enchanting expression style than MS COCO to arouse greater customer interests.
Sentences like ``pearly'', ``so-simple yet so-chic'', ``retro flair'' are more attractive than the plain or ``undecorated'' MS COCO descriptions. 

The image captioning problem has been widely studied and achieved great progress in recent years.
An encoder-decoder paradigm is generally followed with a deep convolutional neural network (CNN)  to encode the input images and a Long Short Term Memory (LSTM) decoder to generate the descriptions~\cite{Kelvin2015,Karpathy2017,Johnson16,Simao2019,Anderson2018}.
The encoder-decoder model is trained via maximum likelihood estimation (MLE), which aims to maximize the likelihood of the next word given the previous words.
However, MLE-based methods will cause the model to generate ``unmatched'' descriptions for the fashion items, where sentences cannot precisely describe the attributes of items. This is due to two reasons.
First, MLE treats the attribute and non-attribute words equally.  Attribute words are not emphasized and directly optimized in the training process, however, they are more important and should be considered as the key parts in the evaluation.
Second, MLE maximizes its objective word-by-word without considering the global semantic meaning of the sentence. This shortcoming may lead to generating a caption that wrongly describes the category of the item. 

To generate better descriptions for fashion items, we propose two semantic rewards as the objective to optimize and train our model using Reinforcement Learning (RL). 
Specifically, we propose an \textit{attribute-level semantic} (ALS) reward with an attribute-matching algorithm to measure the consistency level of attributes between the generated sentences and ground-truth. By incorporating the semantic metric of attributes into our objective, we increase the quality of sentence generation from the semantic perspective.
As a second procedure, we propose a \textit{sentence-level semantic} (SLS) reward to capture the semantic meaning of the whole sentence.
Given a text classifier pretrained on the sentence category classification task, the high level features of the generated description, i.e., the category feature, should stay the same as the ground-truth sentence.
In this paper, we use the output probability of the generated sentence as the groundtruth category as the SLS reward.
Since both ALS reward and SLS reward are non-differentiable, we seek RL to optimize them.

In addition, to guarantee that the image features extracted from the CNN encoder are meaningful and correct, we design a visual attribute predictor to make sure that the predicted attributes match the ground-truth ones.
Then attributes extracted are used as the condition in the LSTM decoder to produce the words of description. 
This work has three main contributions. 
\begin{enumerate}
    \item We build a large-scale fashion captioning dataset FACAD of over $993$K images which are comprehensively annotated with categories, attributes and descriptions.  To the best of our knowledge, it is the first fashion captioning dataset available. We expect that this dataset will greatly benefit the research community, in not only developing various fashion related algorithms and applications, but also helping visual language related studies.  
    \item We introduce two novel rewards (ALS and SLS) into the Reinforcement Learning framework to capture the semantics at both the attribute level and the sentence level to largely increase the accuracy of fashion captioning.
    \item We introduce a visual attribute predictor to better capture the attributes of the image.
The generated description seeded on the attribute information can more accurately describe the item.
\end{enumerate}

\section{Related Work}
\subsubsection{Fashion Studies}
Most of the fashion related studies~\cite{Cucurull_2019_CVPR,Vasileva18FasionCompatibility,Wang2018AttentiveFG,liuLQWTcvpr16DeepFashion,Yu2018,Liu:2012,2018arXiv180610787G} involve images.
For outfit recommendation, Cucurull \textit{et al.}~\cite{Cucurull_2019_CVPR} used a graph convolutional neural network to model the relations between items in a outfit set,  while Vasileva \textit{et al.}~\cite{Vasileva18FasionCompatibility} used a triplet-net to integrate the type information into the recommendation.
Wang \textit{et al.}~\cite{Wang2018AttentiveFG} used an attentive fashion grammar network for landmark detection and clothing category classification.
Yu \textit{et al.}~\cite{Yu2018} introduced the aesthetic information, which is highly relevant with user preference, into clothing recommending systems. Text information has also been exploited. Han \textit{et al.}~\cite{han2017learning} used title features to regularize the image features learned.
Similar techniques were used in ~\cite{Vasileva18FasionCompatibility}.
But no previous studies focus on fashion captioning.
\subsubsection{Image Captioning}
Image captioning helps machine understand visual information and express it in natural language, and has attracted increasingly interests in computer vision. State-of-the-art approaches~\cite{Kelvin2015}\cite{Johnson16}\cite{Simao2019}\cite{Anderson2018} mainly use encoder-decoder frameworks with attention to generate captions for images.
Xu \textit{et al.}~\cite{Kelvin2015} developed soft and hard attention mechanisms to focus on different regions in the image when generating different words.
Johnson \textit{et al.}~\cite{Johnson16} proposed a fully convolutional localization network to generate dense regions of interest and use the generated regions to generate captions. 
Similarly, Anderson \textit{et al.}~\cite{Anderson2018} and Ma \textit{et al.}~\cite{ma2017} used an object detector like Faster R-CNN~\cite{Ren15} or Mask R-CNN~\cite{He2017MaskR} to extract regions of interests over which an attention mechanism is defined. 
Regardless of the methods used, image captioning generally describes the contents based on the relative positions and relations of objects in an image. 
Fashion Captioning, however, needs to describe the implicit attributes of the item which cannot be easily localized by object detectors.

Recently, policy-gradient methods for Reinforcement Learning (RL) have been utilized to train deep end-to-end systems directly on non-differentiable metrics~\cite{Williams92}. 
Commonly the output of the inference is applied to normalize the rewards of RL. Ren \textit{et al.}~\cite{Zhou2017} introduced a decision-making framework utilizing a \textit{policy network} and a \textit{value network} to collaboratively generate captions with reward driven by visual-semantic embedding.
Rennie \textit{et al.}~\cite{Rennie2017} used self-critical sequence training for image captioning.
The reward is provided using CIDEr~\cite{Vedantam15cider} metric.
Gao \textit{et al.}~\cite{Gao_2019_CVPR} extended \cite{Rennie2017} by running a $n$-step self-critical training.
The specific metrics used in RL approach are hard to generalize to other applications, and optimizing specific metrics often impact other metrics severely. However, the semantic rewards we introduce are general and effective in improving the quality of caption generation.

\section{The FAshion CAptioning Dataset}
We introduce a new dataset - FAshion CAptioning Dataset (FACAD) - to study captioning for fashion items. 
In this section, we will describe how FACAD is built and what are its special properties.
\subsection{Data Collection, Labeling and Pre-Processing}
\label{sec:data}
We mainly crawl fashion images with detailed information using Google Chrome, which can be exploited for the fashion captioning task.
Each clothing item has on average $6\sim7$ images of various colors and poses.
The resolution of the images is $1560\times 2392$, much higher than other fashion datasets.

\begin{figure}
\centering
\begin{subfigure}{0.39\textwidth}
\centering
  \includegraphics[width=\textwidth]{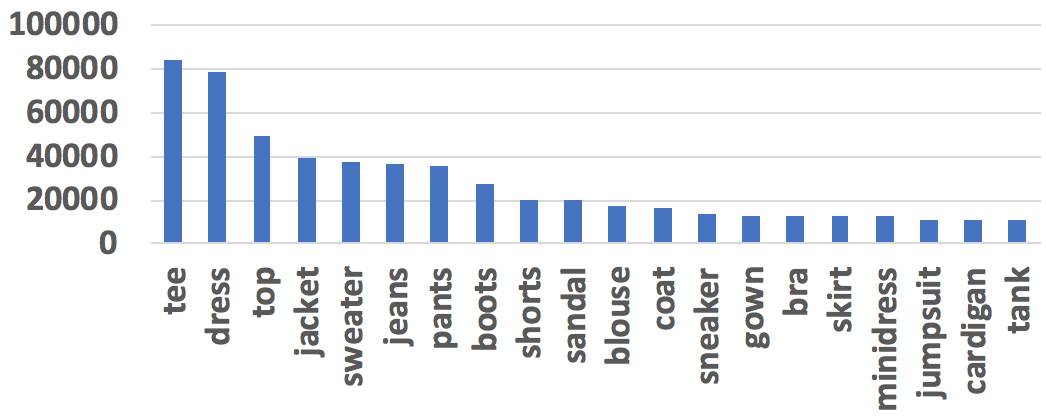}
  \caption{}
  \label{fig:cate} 
\end{subfigure}
\begin{subfigure}{0.60\textwidth}
\centering
  \includegraphics[width=\textwidth]{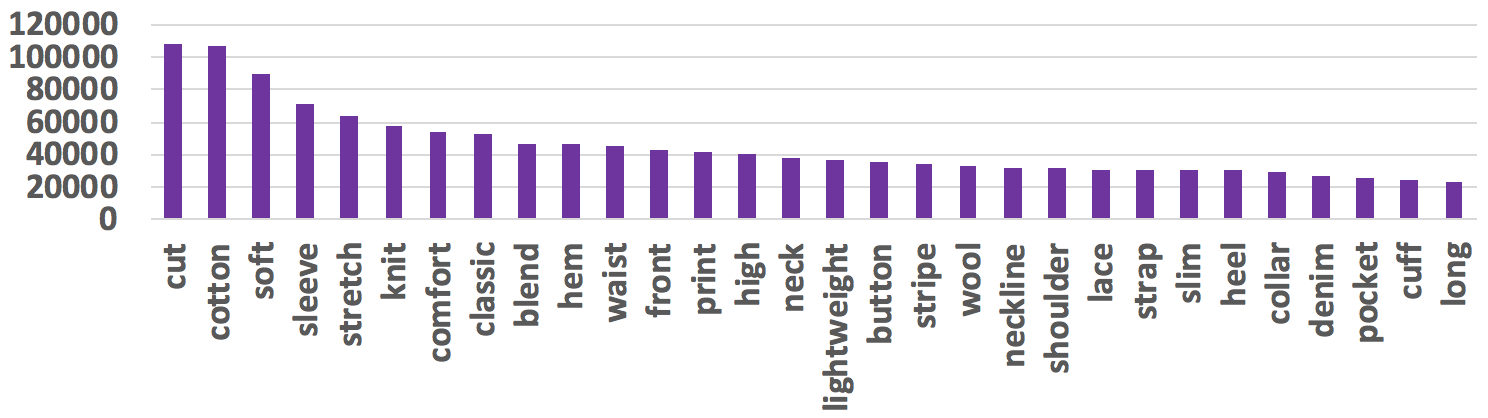}
  \caption{}
  \label{fig:attr}
\end{subfigure}
\caption[]{(a) Number of items in the top-20 categories. (b) Number of items in the top-30 attributes.}
\label{fig:stat}
\end{figure}

In order to better understand fashion items, we label them with rich categories and attributes.
An example category of clothes can be ``dress''  or ``T-shirt'', while an attribute such as ``pink'' or ``lace'' provides some detailed information about a specific item.
The list of the categories is generated by picking the last word of the item titles. 
After manual selection and filtering, there are 472 total valuable categories left. 
We then merge similar categories and only keep ones that contain over 200 items, resulting in 78 unique categories.
Each item belongs to only one category.
The number of items contained by the top-20 categories are shown in Fig.~\ref{fig:cate}.

Since there are a large number of attributes and each image can have several attributes, manual labeling is non-trivial. 
We utilize the title, description and meta data to help label attributes for the items. 
Specifically, we first extract the nouns and adjectives in the title using Stanford Parser~\cite{socher2013parsing}, and then select a noun or adjective as the attribute word if it also appears in the caption and meta data. 
The total number of attributes we extracted is over 3000 and we only keep those that appear in more than 10 items, resulting in a list of 990 attributes.
Each item owns approximately $7.3$ attributes.
We show the number of items that are associated with the top-30 attributes in Fig.\ref{fig:attr}. 

To have clean captions, we tokenize the descriptions using NLTK tokenizer\footnote{https://www.nltk.org/api/nltk.tokenize.html} and remove the non-alphanumeric words. We lowercase all caption words.

\subsection{Comparison with other datasets}
The statistics of our FACAD is shown in Table~\ref{tab:datasets}. 
Compared with other fashion datasets such as~\cite{liuLQWTcvpr16DeepFashion,DeepFashion2,Zheng18,Zou19,guo2019fashion}, FACAD has two outstanding properties. 
First, it is the biggest fashion datasets, with over $993$K diverse fashion images of all four seasons, ages (kids and adults), categories (clothing, shoes, bag, accessories, etc.), angles of human body (front, back, side, etc.).
Second, it is the first dataset to tackle captioning problem for fashion items. 
130K descriptions with average length of 21 words was pre-processed for future researches. 

Compared with MS COCO~\cite{ChenCOCO15} image captioning dataset, FACAD is different in three aspects. 
First, FACAD contains the fine-grained descriptions of attributes of fashion-related items, while MS COCO  narrates the objects and their relations in general images. 
Second, FACAD has longer captions (21 words per sentence on average) compared with 10.4 words per sentence of the MS COCO caption dataset, imposing more difficulty for text generation. 
Third, the expression style of FACAD is enchanting, while that of MS COCO is plain without rich expressions. 
As illustrated in Fig.~\ref{fig:sample}, words like “pearly”, “so-simple yet so-chic”, “retro flair” are more attractive than the plain MS COCO descriptions, like ``a person in a dress''. 
This special enchanting style is important in better describing an item and attracting more customers, but also imposes another challenge for building the caption models.

\begin{table}[!t]
\caption{Comparison of different datasets. $\ast$ Image sizes are approximate values. CAT: category, AT: attribute, CAP: caption, FC: fashion captioning, IC: image captioning, CLS: fashion classification, SEG: segmentation, RET: retrieval.}
\centering
\begin{tabular}{ccccccccccl}
\toprule
Datasets & \# img & img size$\ast$ & \# CAT  & \# AT & \# CAP & avg len & style & task \\
\midrule
\textbf{FACAD} & \textbf{993K} & \textbf{1560$\times$2392} & \textbf{78} & \textbf{990} & \textbf{130K} & \textbf{21}  & \textbf{enchanting} & \textbf{FC} \\
\hline
MS COCO~\cite{ChenCOCO15} & 123K & 640$\times$480 & -- & -- & 616K & 10.4 & plain & IC  \\
VG~\cite{Krishna17} & 108K & 500$\times$500 & -- & -- & 5040K & 5.7 & plain & IC \\
\hline
DFashion~\cite{liuLQWTcvpr16DeepFashion}\cite{DeepFashion2} & 800K & 700$\times$1000 & 50 & 1000 & -- & -- & -- & CLS \\
Moda~\cite{Zheng18} & 55K & -- & 13 & -- & -- & -- & -- & SEG \\
Fashion AI~\cite{Zou19} & 357K & 512$\times$512 & 6 & 41 & -- & -- & -- & CLS  \\
Fashion IQ~\cite{guo2019fashion} & 77K & 300$\times$400 & 3 & 1000 & -- & -- & -- & RET \\
\bottomrule
\end{tabular}
\label{tab:datasets}
\end{table}

\section{Respecting Semantics for Fashion Captioning}
In this section, we first formulate the basic fashion captioning problem and its general solution using Maximum Likelihood Estimation (MLE).
We then propose a set of strategies to increase the performance of fashion captions: 1)  learning specific fashion attributes from the image; 2) establishing attribute-level and sentence-level semantic rewards so that the caption can be generated to be more similar to the ground truth through Reinforcement Learning (RL); 3) alternative training with MLE and RL to optimize the model.

\subsection{Basic Problem Formulation}
We define a dataset of image-sentence pairs as $\mathcal{D}=\{ (X,Y) \}$.
Given an item image $X$, the objective of Fashion Captioning is to generate a description $Y=\{ y_1,\ldots ,y_T \}$ with a sequence of $T$ words, $y_i \in V^K$ being the $i$-th word, $V^K$ being the vocabulary of $K$ words. The beginning of each sentence is marked with a special $<$BOS$>$ token, and the end with an $<$EOS$>$ token.
We denote $\mathbf{y}_i$ as the embedding for word $y_i$. 
To generate a caption, the objective of our model is to minimize the negative log-likelihood of the correct caption using maximum likelihood estimation (MLE): 
\begin{equation}
    \mathcal{L}_{MLE} = -\sum_{t=1}^{T}\log p(y_t\vert y_{1:t-1}, X)
\label{eq:loglikelihood}.
\end{equation}

\begin{figure}
\centering
\includegraphics[width=0.6\textwidth]{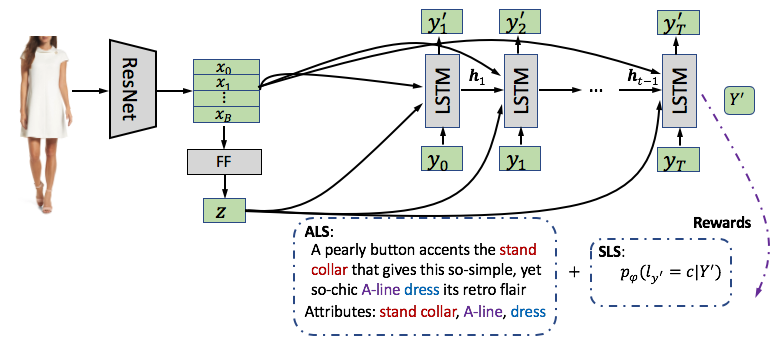}
\caption{The proposed model architecture and rewards.}
\label{fig:model}
\end{figure}

As shown in Fig.~\ref{fig:model}, we use an encoder-decoder architecture to achieve this objective. The encoder is a pre-trained CNN, which takes an image as the input and extracts $B$ image features, $\mathbf{X}=\{ \mathbf{x}_1,\ldots , \mathbf{x}_B \}$.
We dynamically re-weight the input image features $\mathbf{X}$ with an attention matrix $\mathbf{\gamma}$ to focus on specific regions of the image at each time step $t$~\cite{Kelvin2015}, which results in a weighted image feature $\mathbf{x}_t=\sum_{i=1}^B\gamma_t^i\mathbf{x}_i$.
The weighted image feature is then fed into a decoder which is a Long Short-Term Memory (LSTM) network for sentence generation. 
The decoder predicts one word at a time and controls the fluency of the generated sentence. 
More specifically, when predicting the word at the $t$-th step, the decoder takes as input the embedding of the generated word ${y}_{t-1}$, the weighted image feature $\mathbf{x}_t$ and the previous hidden state $\mathbf{h}_{t-1}$. 
The initial memory state and hidden state of the LSTM are initialized by an average of the image features fed through two feed-forward networks $f_c$ and $f_h$ which are trained together with the whole model: $\mathbf{c}_0=f_c(\frac{1}{B}\sum_{i=1}^B\mathbf{x}_i)$, $\mathbf{h}_0=f_h(\frac{1}{B}\sum_{i=1}^B\mathbf{x}_i)$. 
The decoder then outputs a hidden state $\mathbf{h}_{t}$ (Eq.~\ref{eq:hid}) and applies a linear layer $f$ and a \textit{softmax} layer to get the probability of the next word (Eq.~\ref{eq:y}):
\begin{equation}
    \mathbf{h}_{t} = LSTM([\mathbf{y}_{t-1}; \mathbf{x}_t], \mathbf{h}_{t-1})
    \label{eq:hid}
\end{equation}
\begin{equation}
    p_{\theta}(y_t\vert y_{1:t-1}, \mathbf{x}_t)=softmax(f(\mathbf{h}_{t}))
    \label{eq:y}
\end{equation}
where $[;]$ denotes vector concatenation.

\subsection{Attribute Embedding}
To make sure that the caption correctly describes the item attributes, we introduce an attribute feature $\mathbf{z}$ into the model, which modifies Eq.~\ref{eq:loglikelihood} into:
\begin{equation}
    \mathcal{L}_{MLE} = -\sum_{t=1}^{T}\log p(y_t\vert y_{1:t-1}, \mathbf{z}, X)
\label{eq:loglikelihood_z}.
\end{equation}
This objective aims at seeding sentence generation with the attribute feature of the image. To regularize the encoder to output attribute-correct features, we add a visual attribute predictor to the encoder-decoder model.
As each item in the FACAD has its attributes shown in the captions, the predictor can be trained by solving the problem of multi-label classification. 
The trained model can be applied to extract the attributes of an image to produce the caption. 

Fig.~\ref{fig:model} illustrates the attribute prediction network. 
We attach a feed-forward (FF) network to the CNN feature extractor, and its output is fed into a sigmoid layer to produce a probability vector and calculate multi-class multi-label loss.
We can then modify Eq.~\ref{eq:hid} and Eq.~\ref{eq:y} to include the attribute embedding as:
\begin{equation}
    \mathbf{h}_{t} = LSTM([\mathbf{y}_{t-1}; \mathbf{x}_t; \mathbf{z}],  \mathbf{h}_{t-1})
    \label{eq:hid_z}
\end{equation}
\begin{equation}
    p_{\theta}(y_t\vert y_{1:t-1}, \mathbf{x}_t, \mathbf{z})=softmax(f_h(\mathbf{h}_{t}))
    \label{eq:y_z}
\end{equation}
where $\mathbf{z}$ is the attribute features before the output layer, $[;]$ denotes vector concatenation.

\subsection{Increasing the Accuracy of Captioning with Semantic Rewards}
Simply training with MLE can force the model to generate most likely words in the vocabulary, but not help decode the attributes that are crucial to the fashion captioning.
To solve this issue, we propose to exploit two semantic metrics to increase the accuracy of fashion captioning:
an attribute-level semantic reward to encourage our model to generate a sentence with more attributes in the image, and a sentence-level semantic reward to encourage the generated sentence to more accurately describe the category of a fashion item.
Because optimizing the two rewards is a non-differentiable process, during the MLE training, we supplement fashion captioning with a Reinforcement Learning (RL) process. 

In the RL process, our encoder-decoder network with attribute predictor can be viewed as an \textit{agent} that interacts with an external \textit{environment} (words and image features) and takes the \textit{action} to predict the next word. 
After each action, the agent updates its internal \textit{state} (cells and hidden states of the LSTM, attention weights, etc).
Upon generating the \textit{end-of-sequence} ($<$EOS$>$) token, the agent observes a \textit{reward} $r$ as a judgement of how good the overall decision is. We have designed two levels of rewards, as defined below:
\subsubsection{Attribute-Level Semantic (ALS) Reward}
We propose the use of {\em attribute-level semantic} (ALS) reward to encourage our model to locally generate as many correct attributes as possible in a caption.
First, we need to represent an attribute with a phrase. We denote a contiguous sequence of $n$ words as an $n$-gram, and we only consider $n=1, 2$ since nearly all the attributes contain 1 or 2 words. 
We call an $n$-gram that contains a correct attribute a tuple $t_n$. That is, a tuple $t_n$ in the generated sentence contains the attribute in the groundtruth sentence and results in an attribute``Match''.
We define the proportion of ``Matching'' for attributes of $n$ words in a generated sentence as: $P(n)=\frac{Match(n)}{H(n)}$,
where $H(n)$ is the total number of $n$-grams contained by a sentence generated. An $n$-gram may or may not contain an attribute. 
For a generated sentence with $M$ words, $H(n)=M+1-n$. The total number of ``Matches'' is defined as:
\begin{equation}
    Match(n)=\sum_{t_n}\min(C_g(t_n), C_r(t_n))
\end{equation}
where $C_g(t_n)$ is the number of times a tuple $t_n$ occurs in the generated sentence, and 
$C_r(t_n)$ is the number of times the same tuple $t_n$ occurs in the groundtruth caption. We use $\min()$ to make sure that the generated sentence does not contain more repeated attributes than the groundtruth.
We then define the ALS reward as:
\begin{equation}
    r_{ALS}=\beta\{ \prod_{n=1}^2P(n) \}^{\frac{1}{n}}
\end{equation}
where $\beta$ is used to penalize short sentences which is defined as:
\begin{equation}
    \beta = \exp\{ \min(0, \frac{l-L}{l}) \}
\end{equation}
where $L$ is the length of the groundtruth and $l$ is the length of the generated sentence. 
When the generated sentence is much shorter than the groundtruth, although the model can decode the correct attributes with a high reward, the sentence may not be expressive with an enchanting style. We thus leverage a penalization factor to discourage this.

\subsubsection{Sentence-Level Semantic (SLS) Reward}
The use of attribute-level semantic score can help generate a sentence with more correct attributes, which thus increases the similarity of the generated sentence with the groundtruth one at the local level. To further increase the similarity between the generated sentence and groundtruth caption at the global level, we consider enforcing a generated sentence to describe an item with the correct category. This design principle is derived based on our observation that items of the same category share many attributes, while those of different categories often have totally different sets of attributes. Thus, a sentence generally contains more correct attributes if it can describe an item with a correct category. 

To achieve the goal, we pretrain a text category classifier $p_{\phi}$, which is a $3$-layer text CNN, using captions as data and their categories  as labels
($\phi$ denotes the parameters of the classifier).
Taking the generated sentence $Y^{\prime}=\{ y_1^{\prime},\ldots , y_T^{\prime} \}$ as inputs, the text category classifier will output a probability distribution $p_{\phi}(l_{Y^{\prime}}\vert Y^{\prime})$, where $l_{Y^{\prime}}$ is the category label for $Y^{\prime}$. 
The sentence-level semantic reward is defined as:
\begin{equation}
    r_{SLS} = p_{\phi}(l_{Y^{\prime}}=c\vert Y^{\prime})
\end{equation}
where $c$ is the target category of the sentence.

\subsubsection{Overall Semantic Rewards}
To encourage our model to improve both the ALS reward and the SLS reward, we use an overall semantic reward which is a weighted sum of the two:
\begin{equation}
    r = \alpha_1r_{ALS} + \alpha_2r_{SLS}
\end{equation}
where $\alpha_1$ and $\alpha_1$ are two hyper-parameters.

\subsubsection{Computing Gradient with REINFORCE}
The goal of RL training is to minimize the negative expected reward:
\begin{equation}
    \mathcal{L}_{r}=-\mathbb{E}_{Y^{\prime}\sim p_{\theta}}[r(Y^{\prime})]
\end{equation}

To compute the gradient $\nabla_{\theta}\mathcal{L}_{r}(\theta)$, we use the REINFORCE algorithm~\cite{Williams92} to calculate the expected gradient of a non-differentiable reward function.
To reduce the variance of the expected rewards, the gradient can be generalized by incorporating a \textit{baseline} $b$:
\begin{equation}
    \nabla_{\theta}\mathcal{L}_{r}(\theta)=-\mathbb{E}_{Y^{\prime} \sim p_{\theta}}[(r(Y^{\prime})-b)\nabla_{\theta}\log p_{\theta}(Y^{\prime})]
\end{equation}

In our experiments, the expected gradient is approximated using $H$ samples from $p_{\theta}$ and the baseline is the average reward of all the $H$ sampled sentences:
\begin{equation}
\begin{aligned}
    \nabla_{\theta}\mathcal{L}_{r}(\theta) \simeq -\frac{1}{H}\sum_{j=1}^H[(r_j(Y_j^{\prime})-b)\nabla_{\theta}\log p_{\theta}(Y_j^{\prime})]
\end{aligned}
\end{equation}
where $b=\frac{1}{H}\sum_{j=1}^Hr(Y^{\prime}_j)$, 
$Y^{\prime}_j\sim p_{\theta}$ is the $j$-th sampled sentence from model $p_{\theta}$ and $r_j(Y_j^{\prime})$ is its corresponding reward.

\subsection{Joint Training of MLE and RL}
In practice, rather than starting RL training from a random policy model, we warm-up our model using MLE and attribute embedding objective till converge.
We then integrate the pre-trained MLE, attribute embedding, and RL into one model to retrain until it converges again, following the overall loss function:
\begin{equation}
    \mathcal{L} = \mathcal{L}_{MLE} + \lambda_1\mathcal{L}_r + \lambda_2\mathcal{L}_a
\end{equation}
with $\lambda_1$ and $\lambda_2$ being two hyper-parameters.
\section{Experiments}

\subsection{Basic Setting}

\noindent\textbf{Dataset and Metrics}
We run all methods over FACAD. It contains 993K images and 130K descriptions, 
and we split the whole dataset, with approximately 794K image-description pairs for training, 99K for validation, and the remaining 100K for test. 
Images for the same item share the same description.
The number of images associated with one item varies, ranging from 2 to 12. 
As several images in FACAD (e.g., clothes shown in different angles)  share the same description, instead of randomly splitting the dataset, we ensure that the  images with the same caption are contained in the same data split.
We lowercase all sentences and discard non-alphanumeric characters. For words in the training set, we keep the ones that appear at least 5 times, making a vocabulary of 15807 words.

For fair and thorough performance measure, we report results under the commonly used metrics for image captioning, including BLEU~\cite{papineni2002}, METEOR~\cite{denkowski2014meteor}, ROUGEL~\cite{lin2004rouge}, CIDEr~\cite{Vedantam15cider}, SPICE~\cite{spice2016}.
In addition, we compare the attributes in the generated captions with those in the test set as ground truth to find the average precision rate for each attribute using mean average precision (mAP). To evaluate whether the generated captions belong to the correct category, we report the category prediction accuracy (ACC).
We pre-train a $3$-layer text CNN~\cite{kim2014} as the category classifier $p_{\phi}$, achieving a classification accuracy of $90\%$ on testset.

\noindent\textbf{Network Architecture}
As shown in Fig.~\ref{fig:model}, we use a ResNet-101~\cite{He2015DeepRL}, pretrained on ImageNet to encode each image feature. 
Since there is a large domain shift from ImageNet to FACAD, we fine tune the conv4\_x and the conv5\_x layers to get better image features. The features output from the final convolutional layer are used to further train over FACAD.
We use LSTM~\cite{hochreiter1997long} as our decoder. 
The input node dimension and the hidden state dimension of LSTM are both set to $512$. 
The word embeddings of size $512$ are uniformly initialized within $[-0.1, 0.1]$.
After testing with several combinations of the hyper-parameters, we set the $\alpha_1=\alpha_2=1$ to assign equal weights to both rewards, and $\lambda_1=\lambda_2=1$ to balance MLE, attribute prediction and RL objectives during training.
The number of samplings in RL training is $H=5$.

\noindent\textbf{Training Details}
All the models are trained according to the following procedure, unless otherwise specified.
We initialize all models by training using MLE objective with cross entropy loss with ADAM~\cite{kingma2015} optimizer at an initial learning rate of $1\times 10^{-4}$.
We anneal the learning rate by a factor of 0.9 every two epochs.
After the model training converges on the MLE objective, if RL training is further needed in a method, we switch to MLE + RL training till another converge. The overall process takes about 4 days on two NVIDIA 1080 Ti GPUs.

\noindent\textbf{Baseline Methods}
To make fair comparisons, we take image captioning models based both on MLE training and training with MLE$+$RL. For all the baselines, we use their published codes to run the model, performing a hyperparameter search based on the original author's guidelines. We follow their own training schemes to train the models.

\textit{MLE-based Methods.}
\textbf{CNN-C}\cite{Aneja18} is a CNN-based image captioning model which uses a masked convolutional decoder for sentence generation. \textbf{SAT}~\cite{Kelvin2015} applies CNN-LSTM with attention, and we use its hard attention method. \textbf{BUTD}~\cite{Anderson2018} combines the bottom-up and the top-down attention, with the bottom-up part containing a set of salient image regions, each is represented by a pooled convolutional feature vector. \textbf{LBPF}~\cite{Qin2019CVPR} uses a look back (LB) approach to introduce attention value from the previous time step into the current attention generation and a predict forward (PF) approach to predict the next two words in one time step. \textbf{TRANS}~\cite{Simao2019} proposes the use of geometric attention for image objects based on Transformer~\cite{Vaswani2017}.

\textit{MLE + RL based Methods.}
\textbf{AC}~\cite{Zhang2017ac} uses actor-critic Reinforcement Learning algorithm to directly optimize on CIDEr metric. \textbf{Embed-RL}~\cite{Zhou2017} utilizes a ``policy'' and a ``value'' network to jointly determine the next best word. 
\textbf{SCST}~\cite{Rennie2017} is a self-critical sequence training algorithm. \textbf{SCNST}~\cite{Gao_2019_CVPR} is a $n$-step self-critical training algorithm extended from~\cite{Rennie2017}. We use 1-2-2-step-maxpro variant which achieved best performance in the paper.

\subsection{Performance Evaluations} 
\noindent\textbf{Results on Fashion Captioning}
Our Semantic Rewards guided Fashion Captioning (SRFC) model achieves the highest scores on all seven metrics.
Specifically, it provides $0.7$, $0.9$, $1.7$, $1.7$, $1.2$, $0.009$ and $0.013$ points of improvement over the best baseline SCNST on BLEU4, METEOR, ROUGEL, CIDEr, SPICE, mAP and ACC respectively, demonstrating the effectiveness of our proposed model in providing fashion captions. 
The improvement mainly comes from 3 parts, attribute embedding training, ALS reward and SLS reward. 
To evaluate how much contribution each part provides to the final results, we remove different components from SRFC and see how the performance degrades.
For SRFC without attribute embedding, our model experiences the performance drops of $0.1$, $0.2$, $0.3$, $0.5$, $0.5$, $0.002$ and $0.003$ points.
After removing ALS, the performance of SRFC drops $0.3$, $0.7$, $1.2$, $1.4$ and $1.2$ points on the first five metrics.
For the same five metrics, the removing of SLS results in higher performance degradation, which indicates that the global semantic reward plays a more important role in ensuring accurate description generation. 
More interestingly, removing ALS produces a larger drop in mAP, while removing SLS impacts more on ACC. This means that ALS focuses more on producing correct attributes locally, while SLS helps ensure the global semantic accuracy of the generated sentence.
Removing both ALS and SLS leads to a large decrease of the performance on all metrics, which suggests that most of the improvement is gained by the proposed two semantic rewards.
Finally, with the removal of all three components, the performance of our model is similar to that of the baselines without using any proposed techniques. This demonstrates that all three components are necessary to have a good performance on fashion captioning.

\begin{table*}[!t]
\centering
\caption{\textbf{Fashion captioning results -} scores of different baseline models as well as different variants of our proposed method. 
\textbf{A}: attribute embedding learning. 
We highlight the \textbf{best} model in bold.}
\begin{tabular}{ccccccccl}
\toprule
 \textbf{Model} & \textbf{BLEU4} & \textbf{METEOR} & \textbf{ROUGEL} & \textbf{CIDEr} & \textbf{SPICE} & \textbf{mAP} & \textbf{ACC} \\
\midrule
CNN-C~\cite{Aneja18}  & 2.1 & 7.2 & 16.3 & 20.8 & 6.5 & 0.049 & 0.108  \\
SAT~\cite{Kelvin2015}  & 4.3 & 9.5 & 19.9 & 35.2 & 9.8 & 0.056 & 0.124  \\
BUTD~\cite{Anderson2018}  & 4.4 & 9.7 & 19.6 & 36.9 & 10.1 & 0.058 & 0.127  \\
LBPF~\cite{Qin2019CVPR}  & 4.5 & 9.5 & 19.1 & 36.4 & 9.6 & 0.055 & 0.129  \\
TRANS~\cite{Simao2019}  & 4.2 & 10.2 & 19.9 & 36.7 & 9.9 & 0.061 & 0.131  \\
\midrule
AC~\cite{Zhang2017ac}  & 4.8 & 10.9 & 20.6 & 37.8 & 10.7 & 0.068 & 0.142  \\
Embed-RL~\cite{Zhou2017}  & 5.4 & 11.5 & 21.5 & 38.7 & 10.7 & 0.073 & 0.154  \\
SCST~\cite{Rennie2017}  & 5.6 & 11.8 & 22.0 & 39.7 & 11.6 & 0.080 & 0.162  \\
SCNST~\cite{Gao_2019_CVPR} & 6.1 & 12.3 & 22.5 & 40.4 & 12.2 & 0.086 & 0.169  \\
\midrule
SRFC  & \textbf{6.8} & \textbf{13.2} & \textbf{24.2} & \textbf{42.1} & \textbf{13.4} & \textbf{0.095} & \textbf{0.182}  \\
SRFC$-$A  & 6.7 & 13.0 & 23.9 & 41.6 & 12.9 & 0.093 & 0.179   \\
SRFC$-$ALS  & 6.5 & 12.5 & 23.0 & 40.7 & 12.2 & 0.090 & 0.171   \\
SRFC$-$SLS  & 6.1 & 11.8 & 22.2 & 38.9 & 10.6 & 0.083 & 0.162   \\
SRFC$-$ALS$-$SLS & 5.2 & 10.3 & 20.6 & 37.5 & 10.1 & 0.069 & 0.144 \\
SRFC$-$A$-$ALS$-$SLS & 4.4 & 9.8 & 20.2 & 35.6 & 9.9 & 0.058 & 0.127 \\
\bottomrule
\end{tabular}
\label{tab:quantitative}
\end{table*}

\noindent\textbf{Results with Subjective Evaluation}
As fashion captioning is used for online shopping systems, attracting customers is a very important goal.
Automatically evaluating the ability to attract customers is infeasible.
Thus, we perform human evaluation on the attraction of generated captions from different models.
5 human judges of different genders and age groups are presented with 200 samples each. 
Among five participants, two are below 30, two are from 40 to 50 years old, one is over 60. 
They all have online shopping experiences. 
Each sample contains an image, 10 generated captions from all 10 models, with the sequence randomly shuffled. 
Then they are asked to choose the most attractive caption for each sample.
To show the agreement rate, we calculate Fleiss' kappa based on our existing experimental results, with the rate is in the range of [0.6,0.8] indicating consistent agreement, while the range [0.4,0.6] showing moderate agreement. The agreement rates for different models are SRFC (ours) (0.63), SCNST (0.61), SCST (0.62), Embed-RL (0.54), AC (0.56), TRANS (0.52), LBPF (0.55), BUTD (0.53), SAT (0.55), CNN-C (0.54). 
The results in Table~\ref{tab:human} show that our model produces the most attractive captioning.

\begin{table*}[!t]
\centering
\caption{\textbf{Human evaluation on captioning attraction.} We highlight the \textbf{best} model in bold.}
\begin{tabular}{ccccccccccl}
\toprule
 \textbf{Model} & CNN-C & SAT & BUTD & LBPF & TRANS & AC & Embed-RL & SCST & SCNST & SRFC \\
\midrule
\% best & 7.7 & 7.9 & 8.1 & 10.0 & 8.8 & 8.4 & 8.5 & 10.2 & 10.7 & \textbf{19.7} \\
\bottomrule
\end{tabular}
\label{tab:human}
\end{table*}

\begin{figure}
\centering
\includegraphics[width=0.8\textwidth]{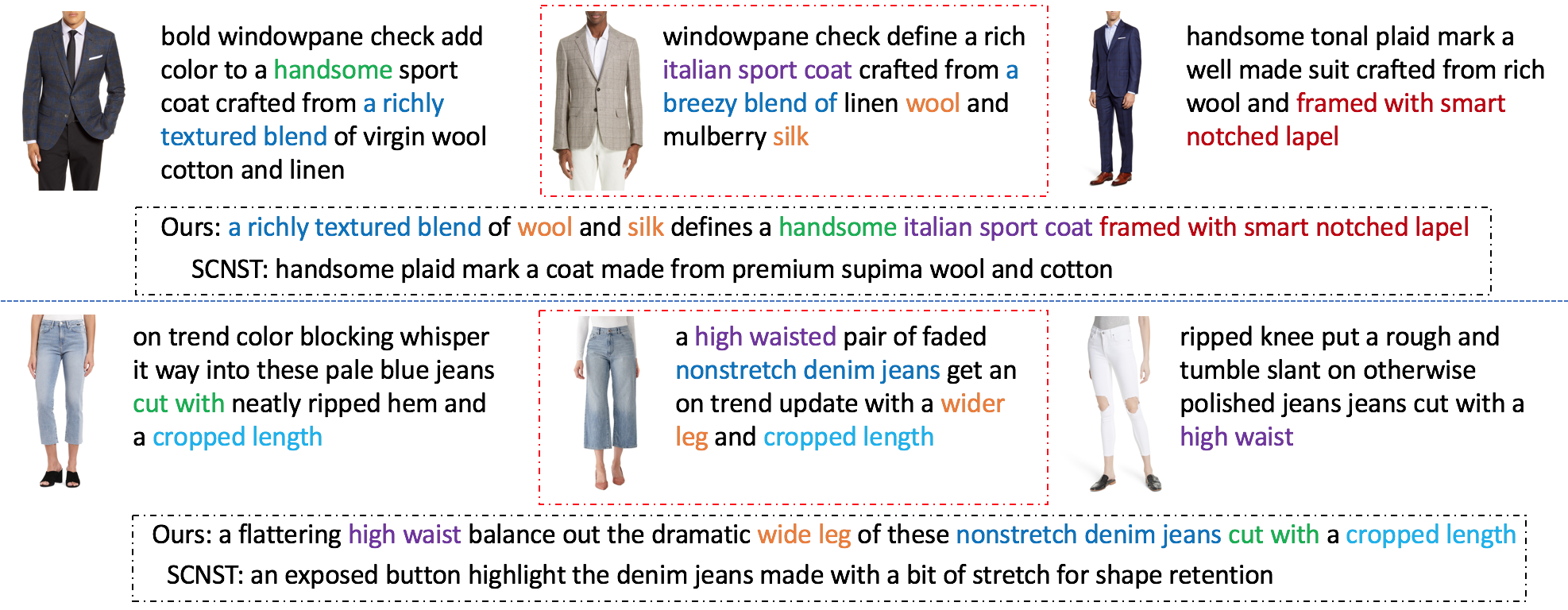}
\caption{Two qualitative results of SRFC compared with the groundtruth and SCNST. 
Two target items and their corresponding groundtruth are shown in the red dash-dotted boxes in the middle column. 
The black dash-dotted boxes contain the captions generated by our model and SCNST.
Our model diversely learns different expressions from the other items (on the first and third columns) to describe the target item.
}
\label{fig:qualitative}
\end{figure}

\noindent\textbf{Qualitative Results and Analysis} Fig.~\ref{fig:qualitative} shows two qualitative results of our model against SCNST and ground truth. In general, our model can generate more reasonable descriptions compared with SCNST for the target image in the middle column. In the first example, we can see that our model generates a description with more details than SCNST, which only correctly predicted the category and some attributes of the target item.

By providing two other items of the same category and their corresponding captions, we have two interesting observations.
First, our model generates descriptions in two steps, it starts learning valuable expressions from similar items (in the same category) based on attributes extracted, and then applies these expressions to describe the target one.
Taking the first item (top row of Fig. \ref{fig:qualitative}) as an example, our model first gets the correct attributes of the image, i.e., \textit{italian sport coat}, \textit{wool}, \textit{silk}.
Then it tries to complete a diverse description by learning from the captions of those items with similar attributes. Specifically, it uses \textit{a richly textured blend} and \textit{handsome} from the first item (left column) and \textit{framed with smart notched lapel} (right column) from the second item to make a new description for the target image. 
The second observation is that our model can enrich description generation by focusing on the attributes identified even if they are not presented in the groundtrue caption.
Even though the \textit{notched lapel} is not described by the ground-truth caption, our model correctly discovers this attribute and generates \textit{framed with smart notched lapel} for it. 
This is because that \textit{notched lapel} is a frequently referred attribute for items of the category \textit{coat}, and this attribute appears in $11.4\%$ descriptions.
Similar phenomena can be found for the second result.
The capability of extracting the correct attributes owes to the \textit{Attribute Embedding Learning} and \textit{ALS} modules.
The \textit{SLS} can help our model generate diverse captions by referring to those from other items with the same category and similar attributes.

\section{Conclusion}
In this work, we propose a novel learning framework for \textit{fashion captioning} and create the first fashion captioning dataset FACAD.
In light of describing fashion items in a correct and expressive manner, we define two novel metrics ALS and SLS, based on which we concurrently train our model with  MLE, attribute embedding and RL training. 
Since this is the first work on fashion captioning, we apply the evaluation metrics commonly used in the general image captioning.
Further research is needed to develop better evaluation metrics.

\section{Acknowledgements}
This work is supported in part by the National Science Foundation under Grants NSF ECCS 1731238 and NSF CCF 2007313.

\clearpage
%
%
\bibliographystyle{splncs04}
\bibliography{egbib}

\begin{thebibliography}{10}
\providecommand{\url}[1]{\texttt{#1}}
\providecommand{\urlprefix}{URL }
\providecommand{\doi}[1]{https://doi.org/#1}

\bibitem{spice2016}
Anderson, P., Fernando, B., Johnson, M., Gould, S.: Spice: Semantic
  propositional image caption evaluation. In: ECCV (2016)

\bibitem{Anderson2018}
Anderson, P., He, X., Buehler, C., Teney, D., Johnson, M., Gould, S., Zhang,
  L.: Bottom-up and top-down attention for image captioning and visual question
  answering. In: The IEEE Conference on Computer Vision and Pattern Recognition
  (CVPR) (2018)

\bibitem{Aneja18}
{Aneja}, J., {Deshpande}, A., {Schwing}, A.G.: Convolutional image captioning.
  In: 2018 IEEE/CVF Conference on Computer Vision and Pattern Recognition
  (2018)

\bibitem{ChenCOCO15}
Chen, X., Fang, H., Lin, T.Y., Vedantam, R., Gupta, S., Dollár, P., Zitnick,
  C.L.: Microsoft coco captions: Data collection and evaluation server.  (2015)

\bibitem{Cucurull_2019_CVPR}
Cucurull, G., Taslakian, P., Vazquez, D.: Context-aware visual compatibility
  prediction. In: The IEEE Conference on Computer Vision and Pattern
  Recognition (CVPR) (June 2019)

\bibitem{denkowski2014meteor}
Denkowski, M., Lavie, A.: Meteor universal: Language specific translation
  evaluation for any target language. In: Proceedings of the Ninth Workshop on
  Statistical Machine Translation (2014)

\bibitem{2018arXiv180610787G}
{Gabale}, V., {Prabhu Subramanian}, A.: {How To Extract Fashion Trends From
  Social Media? A Robust Object Detector With Support For Unsupervised
  Learning}. ArXiv e-prints  (2018)

\bibitem{Gao_2019_CVPR}
Gao, J., Wang, S., Wang, S., Ma, S., Gao, W.: Self-critical n-step training for
  image captioning. In: The IEEE Conference on Computer Vision and Pattern
  Recognition (CVPR) (2019)

\bibitem{DeepFashion2}
Ge, Y., Zhang, R., Wu, L., Wang, X., Tang, X., Luo, P.: A versatile benchmark
  for detection, pose estimation, segmentation and re-identification of
  clothing images. CVPR  (2019)

\bibitem{guo2019fashion}
Guo, X., Wu, H., Gao, Y., Rennie, S., Feris, R.: The fashion iq dataset:
  Retrieving images by combining side information and relative natural language
  feedback. arXiv preprint arXiv:1905.12794  (2019)

\bibitem{han2017learning}
Han, X., Wu, Z., Jiang, Y.G., Davis, L.S.: Learning fashion compatibility with
  bidirectional lstms. In: ACM Multimedia (2017)

\bibitem{He2017MaskR}
He, K., Gkioxari, G., Doll{\'a}r, P., Girshick, R.B.: Mask r-cnn. 2017 IEEE
  International Conference on Computer Vision (ICCV)  (2017)

\bibitem{He2015DeepRL}
He, K., Zhang, X., Ren, S., Sun, J.: Deep residual learning for image
  recognition. 2016 IEEE Conference on Computer Vision and Pattern Recognition
  (CVPR)  (2016)

\bibitem{He2017RealTimeFC}
He, Y., Yang, L., Chen, L.: Real-time fashion-guided clothing semantic parsing:
  A lightweight multi-scale inception neural network and benchmark. In: AAAI
  Workshops (2017)

\bibitem{Simao2019}
Herdade, S., Kappeler, A., Boakye, K., Soares, J.: Image captioning:
  Transforming objects into words. In: Advances in Neural Information
  Processing Systems 32 (2019)

\bibitem{hochreiter1997long}
Hochreiter, S., Schmidhuber, J.: Long short-term memory. Neural computation
  (1997)

\bibitem{Johnson16}
Johnson, J., Karpathy, A., Fei-Fei, L.: Densecap: Fully convolutional
  localization networks for dense captioning. In: Proceedings of the IEEE
  Conference on Computer Vision and Pattern Recognition (2016)

\bibitem{Karpathy2017}
Karpathy, A., Fei-Fei, L.: Deep visual-semantic alignments for generating image
  descriptions. IEEE Trans. Pattern Anal. Mach. Intell. pp. 664--676 (2017)

\bibitem{kim2014}
Kim, Y.: Convolutional neural networks for sentence classification. In:
  Proceedings of the 2014 Conference on Empirical Methods in Natural Language
  Processing ({EMNLP}) (2014)

\bibitem{kingma2015}
Kingma, D.P., Ba, J.: Adam: A method for stochastic optimization (2015)

\bibitem{Krishna17}
Krishna, R., Zhu, Y., Groth, O., Johnson, J., Hata, K., Kravitz, J., Chen, S.,
  Kalantidis, Y., Li, L., Shamma, D.A., Bernstein, M.S., Li, F.: Visual genome:
  Connecting language and vision using crowdsourced dense image annotations.
  International Journal of Computer Vision  (2017)

\bibitem{lin2004rouge}
Lin, C.Y.: {ROUGE}: A package for automatic evaluation of summaries. In: Text
  Summarization Branches Out (2004)

\bibitem{Liu:2012}
Liu, S., Feng, J., Song, Z., Zhang, T., Lu, H., Xu, C., Yan, S.: Hi, magic
  closet, tell me what to wear! In: Proceedings of the 20th ACM International
  Conference on Multimedia (2012)

\bibitem{liuLQWTcvpr16DeepFashion}
Liu, Z., Luo, P., Qiu, S., Wang, X., Tang, X.: Deepfashion: Powering robust
  clothes recognition and retrieval with rich annotations. In: Proceedings of
  IEEE Conference on Computer Vision and Pattern Recognition (CVPR) (2016)

\bibitem{Lu_2019_CVPR}
Lu, Z., Hu, Y., Jiang, Y., Chen, Y., Zeng, B.: Learning binary code for
  personalized fashion recommendation. In: The IEEE Conference on Computer
  Vision and Pattern Recognition (CVPR) (June 2019)

\bibitem{ma2017}
Ma, C.Y., Kadav, A., Melvin, I., Kira, Z., Alregib, G., Graf, H.: Attend and
  interact: Higher-order object interactions for video understanding  (2017)

\bibitem{papineni2002}
Papineni, K., Roukos, S., Ward, T., Zhu, W.J.: {B}leu: a method for automatic
  evaluation of machine translation. In: Proceedings of the 40th Annual Meeting
  of the Association for Computational Linguistics (2002)

\bibitem{Qin2019CVPR}
Qin, Y., Du, J., Zhang, Y., Lu, H.: Look back and predict forward in image
  captioning. In: The IEEE Conference on Computer Vision and Pattern
  Recognition (CVPR) (2019)

\bibitem{Ren15}
Ren, S., He, K., Girshick, R., Sun, J.: Faster r-cnn: Towards real-time object
  detection with region proposal networks. In: Advances in Neural Information
  Processing Systems 28 (2015)

\bibitem{Zhou2017}
Ren, Z., Wang, X., Zhang, N., Lv, X., Li, L.J.: Deep reinforcement
  learning-based image captioning with embedding reward  (2017)

\bibitem{Rennie2017}
Rennie, S.J., Marcheret, E., Mroueh, Y., Ross, J., Goel, V.: Self-critical
  sequence training for image captioning. IEEE Conference on Computer Vision
  and Pattern Recognition (CVPR)  (2017)

\bibitem{socher2013parsing}
Socher, R., Bauer, J., Manning, C.D., Ng, A.Y.: Parsing with compositional
  vector grammars. In: Proceedings of the 51st Annual Meeting of the
  Association for Computational Linguistics (Volume 1: Long Papers) (2013)

\bibitem{Vasileva18FasionCompatibility}
Vasileva, M.I., Plummer, B.A., Dusad, K., Rajpal, S., Kumar, R., Forsyth, D.:
  Learning type-aware embeddings for fashion compatibility. In: ECCV (2018)

\bibitem{Vaswani2017}
Vaswani, A., Shazeer, N., Parmar, N., Uszkoreit, J., Jones, L., Gomez, A.N.,
  Kaiser, L.u., Polosukhin, I.: Attention is all you need. In: Advances in
  Neural Information Processing Systems 30 (2017)

\bibitem{Vedantam15cider}
Vedantam, R., Zitnick, C.L., Parikh, D.: Cider: Consensus-based image
  description evaluation. In: CVPR (2015)

\bibitem{Wang2018AttentiveFG}
Wang, W., Xu, Y., Shen, J., Zhu, S.C.: Attentive fashion grammar network for
  fashion landmark detection and clothing category classification. 2018
  IEEE/CVF Conference on Computer Vision and Pattern Recognition  (2018)

\bibitem{Wang2017ClothingRW}
Wang, Z., Gu, Y., Zhang, Y., Zhou, J., Gu, X.: Clothing retrieval with visual
  attention model. 2017 IEEE Visual Communications and Image Processing (VCIP)
  pp.~1--4 (2017)

\bibitem{Williams92}
Williams, R.J.: Simple statistical gradient-following algorithms for
  connectionist reinforcement learning. Machine Learning  (1992)

\bibitem{Kelvin2015}
Xu, K., Ba, J., Kiros, R., Cho, K., Courville, A., Salakhudinov, R., Zemel, R.,
  Bengio, Y.: Show, attend and tell: Neural image caption generation with
  visual attention. In: Proceedings of the 32nd International Conference on
  Machine Learning (2015)

\bibitem{Yu2018}
Yu, W., Zhang, H., He, X., Chen, X., Xiong, L., Qin, Z.: Aesthetic-based
  clothing recommendation. In: Proceedings of the 2018 World Wide Web
  Conference (2018)

\bibitem{Zhang2017ac}
Zhang, L., Sung, F., Liu, F., Xiang, T., Gong, S., Yang, Y., Hospedales, T.M.:
  Actor-critic sequence training for image captioning. NIPS workshop  (2017)

\bibitem{Zheng18}
Zheng, S., Yang, F., Kiapour, M.H., , Piramuthu., R.: Modanet: A large-scale
  street fashion dataset with polygon annotations. In: ACM Multimedia (2018)

\bibitem{Zou19}
Zou, X., Kong, X., Wong, W., Wang, C., Liu, Y., ang Cao: Fashionai: A
  hierarchical dataset for fashion understanding. In: CVPRW (2019)

\end{thebibliography}
\end{document}